\documentclass{article}

\PassOptionsToPackage{numbers, compress}{natbib}

\usepackage[preprint]{neurips_2020}

\usepackage[utf8]{inputenc} 
\usepackage[T1]{fontenc}    
\usepackage{hyperref}       
\usepackage{url}            
\usepackage{booktabs}       
\usepackage{amsfonts}       
\usepackage{nicefrac}       
\usepackage{microtype}      
\usepackage{color}
\usepackage{booktabs}
\usepackage{graphicx}
\usepackage{booktabs}
\usepackage{multirow}
\usepackage{amsmath}

\title{Automatic Speech Verification Spoofing Detection}

\author{%
  Shentong Mo \\
   Electrical and Computer Engineering \\
  Carnegie Mellon University\\
  Pittsburgh, PA 15213 \\
   \texttt{shentonm@andrew.cmu.edu} \\
   \And
   Haofan Wang \\
   Electrical and Computer Engineering \\
  Carnegie Mellon University\\
  Pittsburgh, PA 15213 \\
   \texttt{haofanw@andrew.cmu.edu} \\
   \AND
   Pinxu Ren \\
  Electrical and Computer Engineering\\
  Carnegie Mellon University\\
  Pittsburgh, PA 15213 \\
  \texttt{pren@andrew.cmu.edu} \\
   \And
   Ta-Chung Chi \\
   Language Technologies Institute \\
  Carnegie Mellon University\\
  Pittsburgh, PA 15213 \\
   \texttt{tachungc@andrew.cmu.edu} \\
}

\begin{document}

\maketitle

\begin{abstract}
Automatic speech verification (ASV) is the technology to determine the identity of a person based on their voice. While being convenient for identity verification, we should aim for the highest system security standard given that it is the safeguard of valuable digital assets. Bearing this in mind, we follow the setup in ASVSpoof 2019 competition to develop potential countermeasures that are robust and efficient. Two metrics, EER and t-DCF, will be used for system evaluation.
\end{abstract}

\section{Introduction}

\begin {figure}[h]
\centering
\includegraphics[width=\columnwidth]{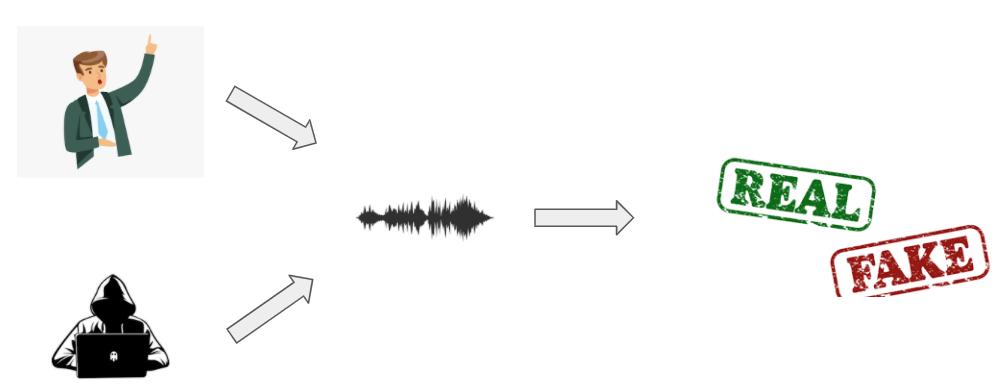}
\caption{Illustration of spoof detection task. The given audio comes from real speaker or synthetic data, the goal is the distinguish whether it is real or not.}
\label{pipeline}
\end {figure}

We are working on the problem of Automatic Speech Verivication Spoofing (ASVSpoof). Roughly speaking, the goal of this task is to distinguish the fake audio input from the real one. The most challenging aspect of this task is that fake audio samples become harder to identify considering the development of neural speech synthesis and voice conversion techniques nowadays. Nevertheless, this is an important task that draws attention from many scholars and research teams, proven by three ASVSpoof competitions hold in 2015 to 2019. There are also practical impacts of ASV spoofing in real world, for example, smart devices such as Alexa or Google Home should not be activated by a random person; Instead, they should only respond to authorized users by recognizing their identities through voice.

We mainly follow the competition rules in 2019 competition \cite{todisco2019asvspoof}. In its essence, the 2019 competition raises two scenarios and three forms of attack that promote the development of countermesaures to spoofing. In terms of scenarios, the first one is called physical access (PA), where speech data is
assumed to be captured by a microphone in a physical, reverberant space. The second one is called logical access (LA), where speech data is
directly injected to the verification system, bypassing the need of a physical microphone. Three forms of attack come along with these scenarios, which are synthetic (Text-to-Speech, TTS), replayed, and converted voice (Voice conversion, VC).

To justify the improvement of algorithms, we adopt the same metrics in the competition, namely, ASV-centric tandem decision cost function (t-DCF) and equal-error-rate (EER) for evaluation.

\section{Related Work}
Most of the previous works  \cite{lai2019assert,cai2019dku,bialobrzeski2019robust,lavrentyeva2019stc,yang2019sjtu,ISpoffing,gomez2019light,alzantot2019deep,jung2019replay,williams2019speech} rely on model ensemble technique, where underlying models are usually deep neural networks and SVM. Among all variants, CNN-based models and residual blocks are the primary choices. They rely on either MFCC, STFT, ivector, and xvector or a combination of them as input features, while there is also a work \cite{jung2019replay} that tries to train the model in an end-to-end fashion by using raw waveform inputs directly. \cite{cai2019dku} relies on data augmentation method by increasing/decreasing the speed of audio samples. Note that there is one paper \cite{EnModel} raises a potential issue of dataset construction, that is, the duration of silence part at the the end of spoofed samples is longer in PA scenario; Models may rely on this clue to make decisions and ignore the actual content.

\begin {figure}[h]
\centering
\includegraphics[width=\columnwidth]{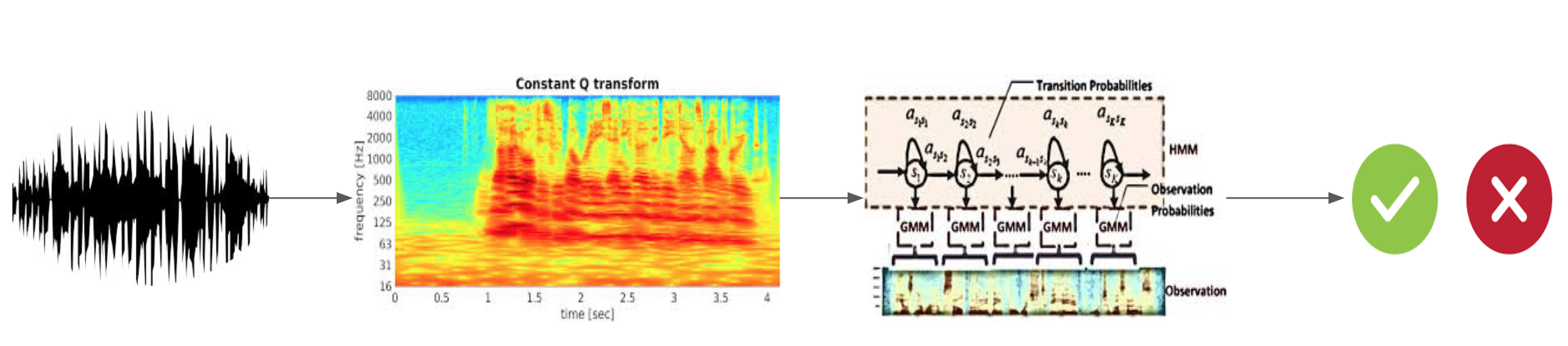}
\caption{Pipeline of ASV sproof detection. Given the raw audio, we first extract the audio features (CQCC in this example), then we feed features into classifier (GMM-HMM) to detect.}
\label{pipeline}
\end {figure}

\section{Datasets}
We are going to use the same dataset as in ASVSpoof 2019 competition. According to \cite{wang2019asvspoof}, it is constructed based on the Voice Cloning Toolkit (VCTK) corpus, which is a multi-speaker English speech database recorded in a hemi-anechoic chamber at a sampling rate of 96 kHz. Given the recorded audio samples, 46 male and 61 female speakers are downsampled to 16 kHz at 16 bits-per-sample. Finally, the set of 107 (46+61) speakers is partitioned into three speaker-disjoint sets for training, development, and evaluation. The training, development, and evaluation sets include 20 training speakers, 10 target and 10 non-target speakers, and 48 target and 19 non-target speakers. 

For fair comparison, we use the same winlen and winstep parameters to match MFCC to our CQCC features. The shape of extracted MFCC features for each audio sample is (time\_step, 13), while the shape of CQCC features should be (time\_step, 60). Both the time steps are the same. 

\section{Methods}
We work on the direction of using multiple models for ensemble. According to our literature survey, most of the previous works use either SVM or some variants of deep model such as CNN and residual blocks. Apart from model architecture, we hypothesize that the most dominating factor should be the quality of extracted features. Currently, we are implementing CQCC, and use other features such as MFCC or raw waveform to augment the expressiveness of input features. Finally, we try to analyze the characteristics of deep model output, maybe TTS models such as wavenet and Tacotron used for generating spoofed samples \cite{wang2020asvspoof} have identifiable output patterns. The pipeline is shown as Figure \ref{pipeline}.

\subsection{Feature Extraction}

\textbf{MFCC feature.}
The MFCC (Mel Frequency Cepstral Coefficients) feature extraction technique basically includes windowing the signal, applying the DFT, taking the log of the magnitude, and then warping the frequencies on a Mel scale, followed by applying the inverse DCT. We adopt the existed package \footnote{https://librosa.org/doc/latest/generated/librosa.feature.mfcc.html} to extract MFCC features in our experiment.

\textbf{CQCC feature.}
CQCC (Constant Q Cepstral Coefficients) is a feature that has proven to be an effective spoofing countermeasure. CQCC features are extracted with the constant Q transform (CQT), a perceptually-inspired alternative to Fourier-based approaches to time-frequency analysis. The extracted mfcc features are of dimension 60.

In our work, we adapt the baseline MATLAB code from ASVSpoof 2019 competition to Python. As there is no usable open source CQCC extractor in Python, we reproduce all functions required by CQCC. We have verified that our CQCC can get similar results as the original MATLAB version.

\subsection{Classifiers}
\textbf{Support Vector Machine (SVM).}
In this project, we are not allowed to use neural network models, so we resort to support vector machine (SVM), which has proven successful in many machine learning tasks. The SVM classifier aims to maximize the margin/decision boundary of samples; It is, therefore, more robust and generalizable. Kernels such as linear, nonlinear and polynomial can also be utilized for efficient feature transformation. We use the implementation from sklearn library\footnote{https://scikit-learn.org/stable/modules/generated/sklearn.svm.SVC.html}. In our work, the radial basis function (RBF) kernel function is adopted.

\textbf{Gaussian Mixture Model (GMM).}
The Gaussian mixture model can be regarded as a model composed of $K$ single Gaussian models. These $K$ sub-models are the hidden variables of the mixture model (hidden variable). Generally speaking, a mixture model can use any probability distribution. The Gaussian mixture model is used here because the Gaussian distribution has good mathematical properties and good calculation performance.

Motivated by \cite{ISpoffing,EnModel}, our classifier consists of $2$ GMM models that were trained from bonafide and spoofed audio samples separately. We utilized the MFCC features extracted from each sample, calculated and concatenated the delta of MFCC, and stack them together as input. During inference, we get a score from each GMM model. We use the sklearn\footnote{https://scikit-learn.org/stable/modules/mixture.html} to implement our GMM model. 

\section{Experiments}

In this section, we conduct experiments to evaluate the effectiveness of our proposed method. First, we extract audio features used in our experiments, namely, MFCC and CQCC. Next, we evaluate the performance of two classifiers (GMM and SVM) given MFCC and CQCC features as inputs. All the models are implemented in Python. The code of our project can be found at: \href{https://github.com/stoneMo/ASVspoof}{https://github.com/stoneMo/ASVspoof}.

\subsection{Evaluation Metrics}
\paragraph{Classification Accuracy (ACC)} 
Accuracy is always used as the metric for binary classification tasks. We calculate the percentage of correct predictions.

\paragraph{Equal Rate Rate (EER).} ASV systems might generate false alarms (classify real to be fake) and miss alarms (classify fake to be true). Usually, when one arises, the other drops. There is a unique point such that these two rates being equal, which is defined to be Equal Rate Rate (EER), or Cross over Error Rate (CER). A system with lower EER is considered more accurate.

\paragraph{tandem-Decision Cost Function (t-DCF).} t-DCF \cite{todisco2019asvspoof} evaluates the performance of ASV and countermeasure (CM) jointly. The detailed descriptions can be found in \cite{todisco2019asvspoof}.

\subsection{Experimental Results}

\begin{table}[htp]
\centering
\begin{tabular}{@{}ccc@{}}
\toprule
SVM & train          & dev   \\ 
\midrule
MFCC            & 0.96/0.05           & 0.94/0.08         \\
CQCC            & 0.90/0.46           & 0.90/0.49       \\
\bottomrule
\end{tabular}
\caption{Results on SVM model. The two numbers in each cell represents accuracy/EER.}
\label{tab:svm}
\end{table}

\begin{table}[htp]
\centering
\begin{tabular}{@{}ccc@{}}
\toprule
GMM   & train & dev \\
\midrule
MFCC  & 0.84/0.12   & 0.83/0.13  \\ 
CQCC  & 0.80/0.44    & 0.79/0.45  \\ 
\bottomrule
\end{tabular}
\caption{Results on GMM model. The two numbers in each cell represents accuracy/EER.}
\label{tab:gmm}
\end{table}

Table \ref{tab:svm} shows the result of SVM. We can see that MFCC gives better performance than CQCC. In the training and testing process, their accuracy is on par with each other. However, CQCC gives a higher EER compared to MFCC. Note that audio samples have variable length, so we pick the first 50 frames of the MFCC/CQCC features and flatten them as the inputs.

The results of GMM are reported in Table  \ref{tab:gmm}. As shown similarly in Table  \ref{tab:svm}, MFCC feature still works better CQCC feature. Although the classification accuracy of GMM is lower than SVM, it shows a lower EER on both training and testing sets. To further boost the performance, we resort to ensemble techniques. Specifically, we adopt the adaboost algorithm with the base algorithm being SVM considering its superior performance. However, it turns out that adaboost does not help with the performance at all. The prediction EER on dev set is only 0.5, which is worse than the numbers reported in Table \ref{tab:svm}. More AdaBoost experiments are left for future work. 

\section{Discussion and Analysis}

\subsection{System Design Limitations}
Ideally, we should use all data instances as inputs. However, due to time and computing constraints, we only sample 1,000 bonafide and spoofed samples for the GMM model.
Obviously, this could hinder the performance. One immediate improvement would be to scale the GMM to use all data instances.

\subsection{Unexpected Results}
\paragraph{MFCC outperforms CQCC.}
According to the original competition paper, CQCC has variable spectrum resolution, and the time-frequency representation is effective in replayed speech detection.
This claim is corroborated by submitted systems in the recent ASVSpoof competition, as most ML-based methods opt for CQCC.
To our surprise, CQCC shows inferior performance in our experiments. Although we tried to be meticulous when re-implementing CQCC (line by line output verification following the official MATLAB code), we might still ignore some subtleties which leads to the degraded performance.

\paragraph{SVM outperforms GMM}
Many methods from ASVSpoof competition use GMM instead of SVM, but our experiment results suggest the opposite. This might be attributed to the system design limitation mentioned above; Due to limited RAM, we are not able to fit the GMM model on all training instances. We have to perform sampling and get a portion of spoofed data for training the spoofed GMM model. With a more powerful machine, we might be able to train the GMM model based on all data instances and derive a different conclusion.

\subsection{Future Works}
Due to time constraint, many issues remain unaddressed. We list them out as they definitely worth further discussion.

\paragraph{Does gender matter?}
We did not conduct individual experiments on male/female's voice samples, so there is not a definite answer for now. However, given the fact that male and female's voice differs in frequency, explicitly modeling the gender difference might help.

\paragraph{The effect of silence}
We notice that silence is present in the voice clips, which may bias the prediction of our models. For example, our models might assert that long silence is spoofed. Our next step will be to reduce the effect of silence by manually removing or equalizing them across samples.

\paragraph{Comparison with baselines}
As per project requirement, we are not allowed to use neural methods, so we only use GMM model as our baseline. In the future, we will implement more competitive baselines as reported in ASVSpoof competition for a fair comparison.

\paragraph{Feature ensemble}
We tried model ensemble by using adaboost with SVM. Similarly, we can perform ensemble in feature space. One possible approach is to train different SVM models with different feature inputs, for example, one with CQCC and the other with MFCC. We haven't seen many methods working on both CQCC and MFCC features, deriving an efficient way to do feature ensemble may shed some light on future research.

\bibliographystyle{plain}
\bibliography{neurips_2020}

\end{document}